\documentclass[sn-mathphys]{sn-jnl}

\usepackage{times}
\usepackage{soul}
\usepackage{comment}
\usepackage{multirow}
\usepackage[utf8]{inputenc}
\usepackage[small]{caption}
\usepackage{graphicx}
\usepackage{amsmath}

\DeclareMathOperator*{\argmin}{arg\,min}

\DeclareMathOperator*{\distance}{Dist}

\usepackage{booktabs}
\usepackage[mathscr]{eucal}

\jyear{2021}%

\theoremstyle{thmstyleone}%
\theoremstyle{thmstyletwo}%

\theoremstyle{thmstylethree}%

\begin{document}

\title[Multi-Center Federated Learning]{Multi-Center Federated Learning: Clients Clustering for Better Personalization}


\author[1]{\fnm{Guodong} \sur{Long}}\email{guodong.long@uts.edu.au}
\equalcont{These authors contributed equally to this work.}

\author[1]{\fnm{Ming} \sur{Xie}}\email{ming.xie@student.uts.edu.au}
\equalcont{These authors contributed equally to this work.}

\author[1]{\fnm{Tao} \sur{Shen}}\email{tao.shen@uts.edu.au}

\author[2,3]{\fnm{Tianyi} \sur{Zhou}}\email{tianyizh@uw.edu}

\author[1]{\fnm{Xianzhi} \sur{Wang}}\email{xianzhi.wang@uts.edu.au}

\author*[1]{\fnm{Jing} \sur{Jiang}}\email{jing.jiang@uts.edu.au}

\affil*[1]{\orgdiv{Australian AI Institute}, \orgname{University of Technology Sydney}, \orgaddress{\country{Australia}}}

\affil[2]{\orgname{University of Washington}, \orgaddress{\city{Seattle}, \state{WA}, \country{USA}}}

\affil[3]{\orgname{University of Maryland}, \orgaddress{\state{MD}, \country{USA}}}


\abstract{Personalized decision-making can be implemented in a Federated learning (FL) framework that can collaboratively train a decision model by extracting knowledge across intelligent clients, e.g. smartphones or enterprises. FL can mitigate the data privacy risk of collaborative training since it merely collects local gradients from users without access to their data. However, FL is fragile in the presence of statistical heterogeneity that is commonly encountered in personalized decision making, e.g., non-IID data over different clients. Existing FL approaches usually update a single global model to capture the shared knowledge of all users by aggregating their gradients, regardless of the discrepancy between their data distributions. By comparison, a mixture of multiple global models could capture the heterogeneity across various clients if assigning the client to different global models (i.e., centers) in FL. To this end, we propose a novel multi-center aggregation mechanism to cluster clients using their models' parameters. It learns multiple global models from data as the cluster centers, and simultaneously derives the optimal matching between users and centers. We then formulate it as an optimization problem that can be efficiently solved by a stochastic expectation maximization (EM) algorithm. Experiments on multiple benchmark datasets of FL show that our method outperforms several popular baseline methods. The experimental source codes are publicly available on the Github repository\footnote{GitHub repository: \url{https://github.com/mingxuts/multi-center-fed-learning}}.}




\maketitle

\section{Introduction}

The widespread of social networks and mobile APPs has witnessed a huge volume of data generated by end-users on mobile devices \cite{cai2020target}. Generally, a service provider on the server side collect users' data and train a global machine learning model such as deep neural networks. Such a centralized machine learning approach causes severe practical issues, e.g., communication costs, consumption of device batteries, and the risk of violating the privacy and of user data. Moreover, each user or client may have different decision making preferences that may caused by the changes of their life events and dynamic networks.

The modern decision-making system is generally empowered by various deep neural network-based intelligent models that include two components: embedding and decision. 
The embedding component could be a backbone is composed of CNN \cite{graziani2020concept,shrestha2021augmenting}, GCN  \cite{xue2021dynamic,li2021deep}, RNN/LSTM \cite{zhang2021episodic,fintz2021using}, or Self-attention/Transformer modules \cite{vaswani2017attention,peng2020self} to transform the multi-modal data, such as images, graphs, time series, sequential behavior and texts, into numeric vectors for further processing.
The decision component usually takes a multi-layer fully connected neural network to model the complex relationship between inputs and decisions (outputs). Then, the embedding and decision components will be connected to form an entire deep learning model to be trained in an end-to-end manner.

Federated learning (FL)~\cite{mcmahan2017communication} is a new machine learning paradigm that learns models collaboratively using the training data distributed on remote devices to boost communication efficiency. 
There are three advantages that can make FL be the best option to implement a personalized decision-making system. 
First, the deep learning model requires a huge volume of training data that is usually impractical for a single client, thus, the FL-based framework enables the clients to collaboratively train a model with abundant training data.
Second, the FL separates the model training and data storage, which can greatly reduce the privacy risk for participating clients.  
Third, each client can take part in the model training process of the FL system, and then it can take the opportunity to train a personalized decision-making model by leveraging both the globally shared knowledge and locally stored data.

The vanilla FL addresses a practical setting of distributed learning, where 1) the central server is not allowed to access any user data which protects users' privacy, and 2) the data distribution over different users is non-IID, which is a natural assumption of real-world applications. However, early FL approaches~\cite{mcmahan2017communication}\cite{yang2019federated} use only one global model as a single-center to aggregate the information of all users. The stochastic gradient descent (SGD) for single-center aggregation is designed for IID data, and therefore, conflicts with the non-IID setting in FL. The observed data in each device are generated or produced by the user. The dataset across devices are usually non-IID that reflects different preferences or decision logic of users. We assume the whole population could be partitioned into different clusters or groups in which the users have similar preferences. Therefore, learning an intelligent model customised for each group with non-IID data can assist users to make decision-making by considering personal preferences.

Recently, the non-IID or heterogeneity challenge of FL has been studied to improve the robustness of global models against outlier/adversarial users and devices~\cite{ghosh2019robust,li2018federated,li2019feddane}. Moreover, ~\cite{sattler2019clustered} proposed an idea of clustered FL (FedCluster) that addresses the non-IID issue by dividing the users into multiple clusters. However, the hierarchical clustering in FedCluster is achieved by multiple rounds of bipartite separation, each requiring the federated SGD algorithm to run until convergence. Hence, its computational and communication efficiency will become bottlenecks when applied to a large-scale FL system. More recently, ~\cite{mansour2020three} and ~\cite{ghosh2020efficient} proposed to cluster the local models according to the loss of hypothesis. In particular, each user will try all $K$ global models representing $K$ clusters, and then select the best global model as the cluster ID by considering the lowest loss of running the global model on local data. However, this posts high communication and computation overheads because the selected nodes will spend more resources for receiving and running multiple global models.

In this paper, we propose a novel multi-center FL framework that updates multiple global models by aggregating information from multiple user groups. In particular, the datasets of the users in the same group are likely to be generated or derived from the same or similar distribution. We formulate the problem of the multi-center FL as the joint clustering of users, and then optimizing of the global model for users in each cluster. In particular, (1) each user's local model is assigned to its closest global model, and (2) the global model in each cluster leads to the smallest loss over all the associated users. 
The proposed multi-center FL not only inherits the communication efficiency of the federated SGD but also retains the capability of handling non-IID data on heterogeneous datasets. 
Lastly, we propose a new optimization method in line with EM algorithm to train our model.

We summarize our main contributions as: 
\begin{itemize}
    \setlength\itemsep{0.3em}
    \item We propose a novel multi-center aggregation approach (Section~\ref{subsec:multi-center}) to address the non-IID challenge of personalized decision-making system.
    \item We design an objective function, namely multi-center federated loss (Section~\ref{subsec:objective}), for user clustering in FL.
    \item We propose Federated Stochastic Expectation Maximization (FeSEM)  (Section~\ref{subsec:algorithm}) to solve the optimization of the proposed objective function.
    \item We present the algorithm as an easy-to-implement and strong baseline for FL. Its effectiveness is evaluated on benchmark datasets. (Section~\ref{sec:experiments})
\end{itemize}

\section{Related work}\label{Relat}
\subsection{Personalized Decision-making}
Decision making is the process of making choices by identifying a decision, gathering information, and assessing alternative resolutions. In most of the scenarios, each individual person usually makes a personal choice given the collected information. To model the personalized decision-making process \cite{mandl2011consumer,pomytkina2020personal}, a general solution is to collect the user's personal characteristics, e.g. demographics \cite{pazzani2007content}, behavior history \cite{schafer2007collaborative}, and social networks \cite{zhang2018dual}, as part of the input to be considered by a centralized intelligent model. This solution usually train a large-scale machine learning or recommendation models at cloud server using the collected personal data from users, thus it will cause privacy concerns. In recent, a new service architecture has been proposed to provide service based on a standalone on-device intelligent \cite{konevcny2016federated} in each smart device. In particular, a unique intelligent model customized for each user will be deployed to the user’s smart device, so as to provide service independently while not relying on the decision from the cloud server. The user's personal data will be stored locally to train the intelligent model, thus no personal data will be uploaded to the server.

\subsection{Federated Learning}
Federated learning (FL) enables users to leverage rich data for machine learning models without compromising their data.
It has attracted a significant amount of research interest since 2017, with many studies investigating FL from several aspects, e.g., system perspective, personalized models, scalability~\cite{Bonawitz2019TowardsFL}, communication efficiency~\cite{Konecn2018FederatedLS}, and privacy~\cite{geyer2017differentially}. Most of the related work addresses a particular concern such as security or privacy~\cite{rouhani2018deepsecure,liu2019revocable,cao2020federated}. It has been applied to various industry applications, such as banking \cite{long2020federated}, healthcare \cite{rieke2020future,xu2021federated,long2022federated}, and mobile internet applications \cite{jiang2020decentralized}.

FL is designed for specific scenarios that can be further expanded to a standard framework to preserve data privacy in large-scale machine learning systems or mobile edge networks \cite{lim2020federated}. For example, \cite{yang2019federated} expanded FL by introducing a comprehensive, secure FL framework that includes horizontal FL, vertical FL, and federated transfer learning. The work in~\cite{li2019federated,lyu2020threats} surveyed the FL systems in relation to their functions on privacy protection and security threats. \cite{kairouz2021advances} discussed the advances and open problems in FL. \cite{caldas2018leaf} proposed LEAF -- a benchmark for federated settings with multiple datasets. \cite{luo2019real} proposed an object detection-based dataset for FL.

Heterogeneity is a core challenge in the federated setting and has been widely studied from various perspectives. \cite{haddadpour2019convergence} conducted theoretical convergence analysis for FL with heterogeneous data. \cite{hsu2019measuring} measured the effects of non-IID data for federated visual classification. \cite{yang2020heterogeneity} proposed a heterogeneity-aware platform design for FL. \cite{liang2020think} discussed the local representations that enable data to be processed on new devices in different ways according to their source modalities instead of using a single global model. The single global model might not generalize to unseen modalities and distributions of data. \cite{li2019fedmd} proposed a new federated setting composed of a shared global dataset and many heterogeneous datasets from devices. \cite{jeong2018communication} and \cite{lin2020ensemble} proposed to integrate knowledge distillation with FL to tackle the model heterogeneity. \cite{yu2020heterogeneous} proposed a general FL framework to align heterogeneous model architectures and functional neurons. \cite{tan2021fedproto} proposed to a prototype learning-based FL framework to tackle the challenges of task heterogeneity across devices. Particularly, it aggregates the prototypes rather than model parameters, and then the communication efficiency is much higher than gradient-based algorithm, such as FedAvg.  

To solve the problem caused by non-IID data in a federated setting \cite{caldas2018leaf}, \cite{sattler2019clustered} proposed clustered FL (FedCluster) by integrating FL and bi-partitioning-based clustering into an overall framework, and \cite{mansour2020three,ghosh2020efficient} proposed a hypothesis-based federated clustering that assigns the cluster by considering the loss of running the global model on local data. \cite{ghosh2019robust} proposed a robust FL comprising three steps: 1) learning a local model on each device, 2) clustering model parameters to multiple groups, each being a homogeneous dataset, and 3) running a robust distributed optimization~\cite{li2019rsa} in each cluster. \cite{ma2022convergence} propose a general form to model the clustered FL problem into a bi-level optimization framework, and then conduct theoretical analysis on the convergence.

\cite{li2019feddane} proposed FedDANE by adapting the DANE~\cite{shamir2014communication} to a federated setting. In particular, FedDANE is a federated Newton-type optimization method. \cite{li2018federated} proposed FedProx for the generalization and re-parameterization of FedAvg~\cite{mcmahan2017communication}. It adds a proximal term to the objective function of each device's supervised learning task, and the proximal term is to measure the parameter-based distance between the server and the local model. \cite{arivazhagan2019federated} added a personalized layer for each local model, i.e., FedPer, to tackle heterogeneous data. \cite{chen2022personalized} propose to a structure-based model aggregation mechanism to enhance the personalized federated learning.

\section{Background}\label{Problem}
\subsection{Overall framework}
\begin{figure}[htbp!] 
    \centering
    \includegraphics[width=0.96\columnwidth]{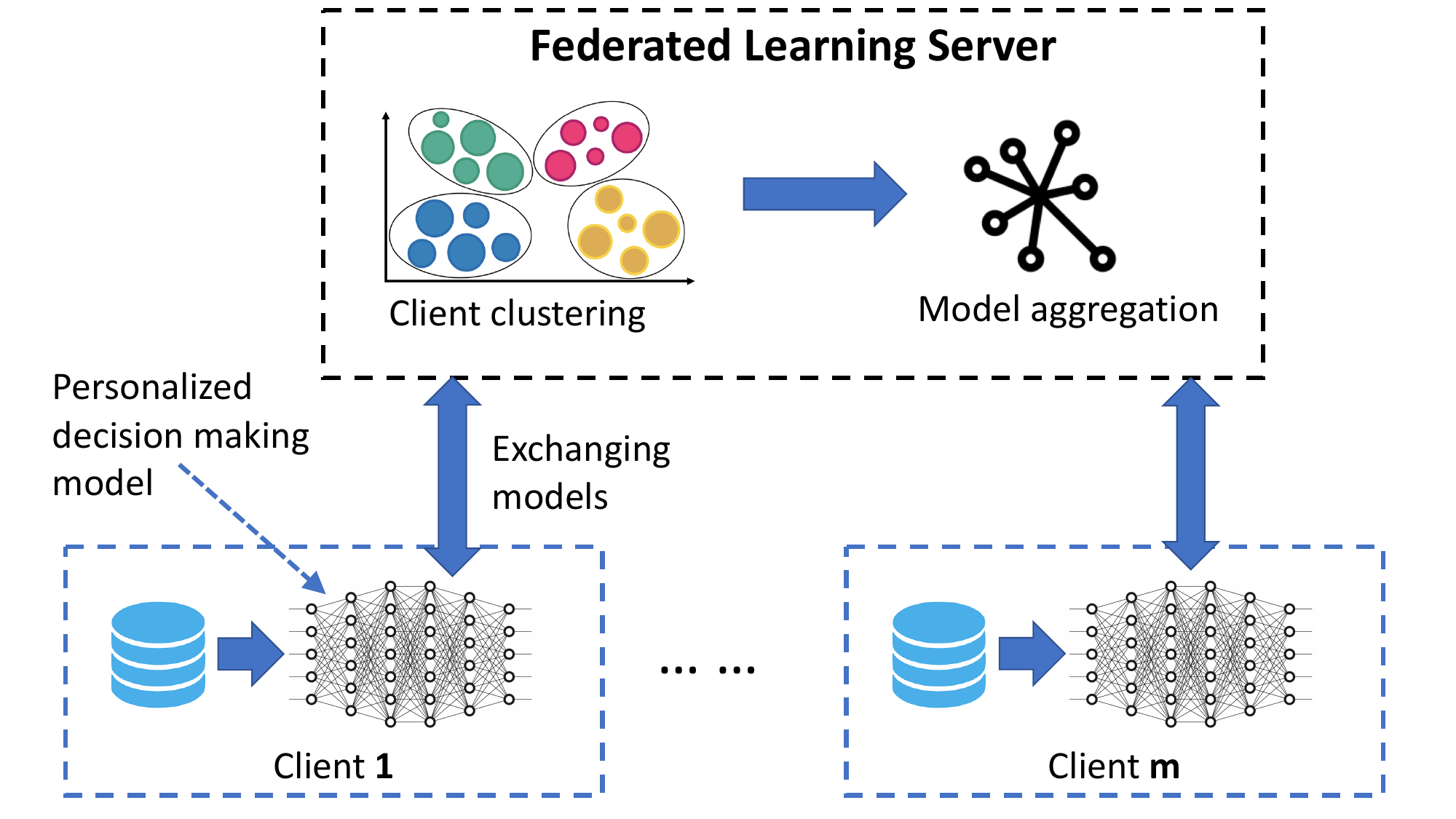}
    \caption{\footnotesize Overall framework of multi-center Federated Learning.
    }
    \label{fig:framework}
\end{figure}

This section will introduce the overall framework of the proposed multi-center Federated Learning. As shown in Fig. \ref{fig:framework}, each client could be an intelligent device or computer in an enterprise, and they will collaboratively train an intelligent model via a coordinating server. First, the client initializes its local model using the global model from the federated learning server. Second, the client trains the local model using its own data. Third, the trained model will be uploaded into the server that will conduct client clustering across clients' model parameters, and then conduct model aggregation in each cluster. Fourth, the model in each cluster will be dispatched to the corresponding clients. The multi-center FL framework will repeat these four steps till convergence or the stop condition is satisfied. 

In the following subsections, we will give a formal formulation of the proposed problem, and then explain the motivation of the design. The detailed method and optimization procedure is provided in Section \ref{sec:method}.

\subsection{Problem Setting}
In FL, each device-$i$ has a private dataset $\mathcal{D}_i=\{\mathcal{X}_i, \mathcal{Y}_i \}$, where $\mathcal{X}_i$ and $\mathcal{Y}_i$ denote the input features and corresponding gold labels respectively. Each dataset $\mathcal{D}_i$ will be used to train a local supervised learning model $\mathcal{M}_i:\mathcal{X}_{i} \rightarrow \mathcal{Y}_{i}$. $\mathcal{M}$ denotes a deep neural model parameterized by weights $W$. It is built to solve a specific task, and all devices share the same model architecture. 

For the $i$-th device, given a private training set $\mathcal{D}_i$, the training procedure of $\mathcal{M}_i$ is represented in brief as
\begin{equation} 
    \label{eq:fed-sp-loss}
    \min_{W_i} L_s(\mathcal{M}_i, \mathcal{D}_i, W_i),
\end{equation}
where $L_s(\cdot)$ is a general definition of the loss function for any supervised learning task, and its arguments are model structure, training data and learnable parameters respectively, and ${W}'$ denotes the parameters after training. In general, the data from one device is insufficient to train a data-driven neural network with satisfactory performance. An FL framework optimizes the local models in a distributed manner and minimizes the loss of the local data on each device. 

Hence, the optimization in vanilla FL over all the local models can be written as
\begin{equation*} 
    \min_{\{W_i\}_{i=1}^m} \alpha_i L_s(\mathcal{M}_i, \mathcal{D}_i, W_i)
\label{eq:sl-lossw}
\end{equation*}
where $m$ denotes the number of devices, and $\alpha_i = \frac{\lvert \mathcal{D}_i \rvert}{\sum_j \lvert \mathcal{D}_j \rvert} $ is an importance weight that is measured by the number of samples on each clients.  

On the server side, the vanilla FL aggregates all local models into a global one $\mathcal{M}_{global}$ which is parameterized by $\tilde{W}^{g}$. In particular, it adopts a weighted average of the local model parameters $[W_i]_{i=1}^m$, i.e.,
\begin{equation}
    \tilde{W}^{g} = \sum_{i=1}^{m} \alpha_i {W}_i,
\end{equation} 
which is the nearest center for all $\{W_i\}_{i=1}^m$ in terms of a weighted L2 distance:
\begin{equation} \label{eq:nc}
     \tilde{W}^{g} \in\argmin_{\tilde W} \sum_{i=1}^{m} \alpha_i \|\tilde{W} - W_i\|_2^2.
\end{equation}

More generally, we can replace the L2 distance in Eq.~\eqref{eq:nc} by other distance metric $\distance(\cdot,\cdot)$ and minimize the difference between the global model and all the local models, i.e.,
\begin{equation} \label{eq.loss}
\min_{\tilde W}\frac{1}{m}\sum_{i=1}^{m} \distance(W_{i}, \tilde{W}).
\end{equation}

The above aims to find a consistent solution across global model and local models. 
Note that a direct macro average is used here regardless of the weight of each device, which treats every device equally. The weights used in Eq.~\eqref{eq:sl-lossw} can easily be incorporated for a micro average.

The divergence $\distance(\cdot,\cdot)$ between the global model and local models plays an essential role in the FL objective. The simple L2 distance for $\distance(\cdot,\cdot)$ does not take into account the fact that two models can be identical under the arbitrary permutation of neurons in each layer.

Hence, the lack of neuron matching may cause misalignment in that two neurons with similar functions and different indexes cannot be aligned across models \cite{yurochkin2019bayesian}. However, the index-based neuron matching in FL \cite{shamir2014communication} is the most widely used method and works well in various real applications. One potential reason for this is that the index-based neuron matching can also slowly align the function of neurons by repeatedly initializing all local models with the same global model. To simplify the description, we will discuss our method for index-based neuron matching.

\begin{figure}[htbp!] 
    \centering
    \includegraphics[width=0.9\columnwidth]{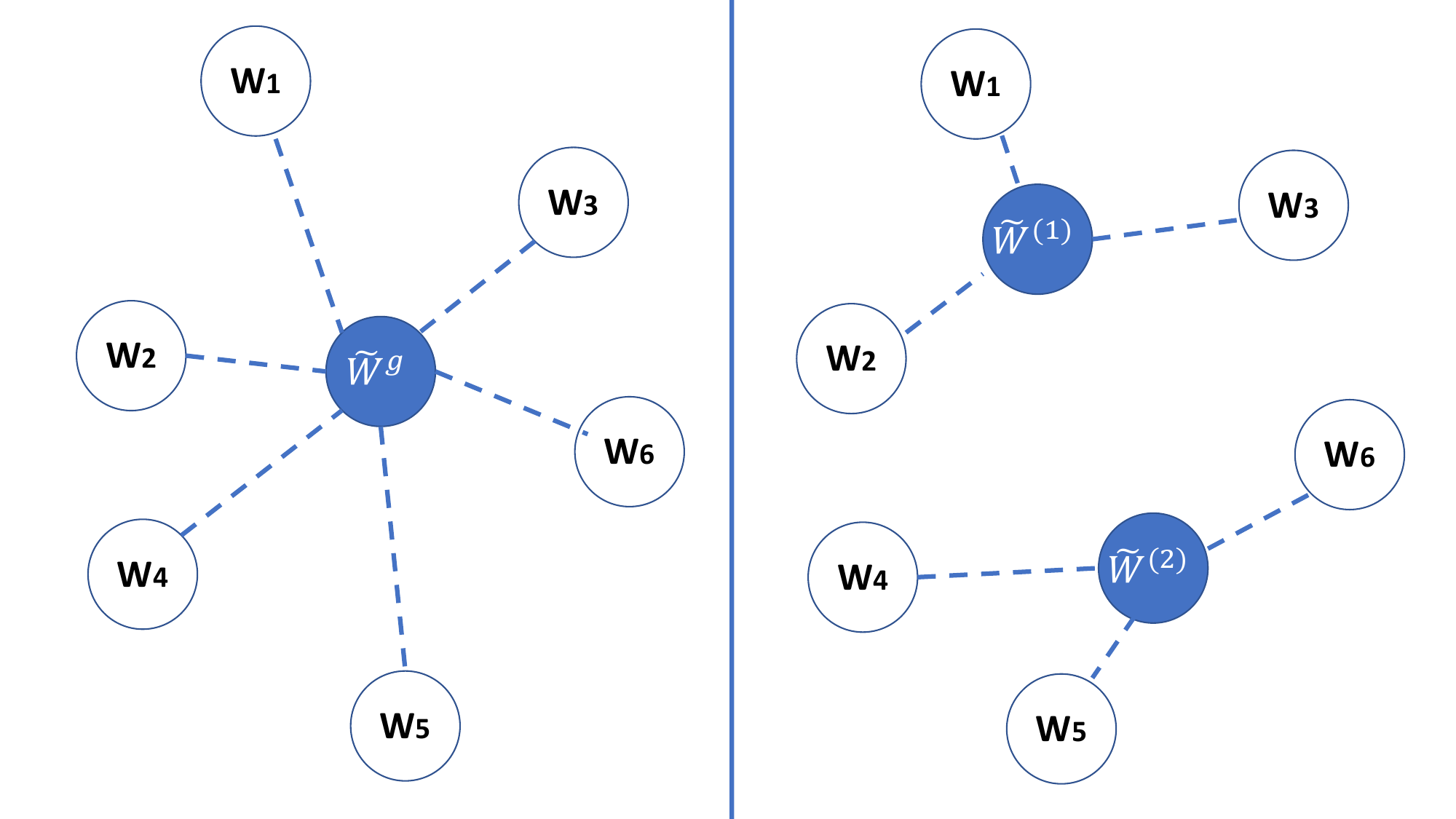}
    \caption{Comparison between single-center aggregation in vanilla FL (left) and multi-center aggregation in the proposed one (right). Each $W_i$ represents the local model's parameters collected from the $i$-th device, which is denoted as a node in the space. $\tilde{W}$ represents the aggregation result of multiple local models.
    }
    \label{fig:aggregation}
\end{figure}

\subsection{Motivation}
Federated learning (FL) usually aggregates all local models to a single global model. However, this single-center aggregation is fragile under heterogeneity. In contrast, we consider FL with multiple centers to better capture the heterogeneity by assigning nodes to different centers so only similar local models are aggregated. 
Consider two extreme cases for the number of centers, $K$: (1) when $K=1$, it reduces to the FedAvg with a single global model, which cannot capture the heterogeneity and the global model might perform poorly on specific nodes; (2) When $K=m$, the heterogeneity problem can be avoided by assigning each node to one global model. But the data on each device used to update each global model can be insufficient and thus we lose the main advantage of FL.
Our goal is to find a sweet point between these two cases to balance the advantages of federated averaging and the degradation caused by underlying heterogeneity.

Learning one unique model for each node has been discussed in some recent FL studies for better personalized models. They focus on making a trade-off between shared knowledge and personalizing. The personalizing strategy either applies fine-tuning of the global model \cite{zhao2018federated} for each node, or only updates a subset of personalized layers for each node \cite{arivazhagan2019federated,liang2020think}, or deploys a regularization term in the objective \cite{deng2020adaptive,t2020personalized,hanzely2020federated}. In contrast, Multi-center FL in this paper mainly focuses to address the heterogeneity challenge by assigning nodes to different global models during aggregation. But it can be easily incorporated in these personalization strategies. 
In the following, we will start from the problem setting for the the vanilla FL, and then elucidate our motivation of improving FL's tolerance to heterogeneity by multi-center design.

\section{Methodology} \label{sec:method}

\subsection{Multi-Center Model Aggregation} \label{subsec:multi-center}
To overcome the challenges arising from the heterogeneity in FL, we propose a novel model aggregation method with multiple centers, each associating with a global model $\tilde{W}^{(k)}$ updated by aggregating a cluster of user's models with nearly IID data. In particular, all the local models will be grouped to $K$ clusters, denoted as $C_1,\cdots,C_K$, each covering a subset of local models with parameters $\{W_j\}_{j=1}^{m_k}$. 

An intuitive comparison between the vanilla FL and our multi-center FL is illustrated in Fig.~\ref{fig:aggregation}. As shown in the left figure, there is only one center model in vanilla FL. In contrast, the multi-center FL shown in the right has two centers, $W^{(1)}$ and $W^{(2)}$, and each center represents a cluster of devices with similar data distributions and models. Obviously, the right one has a smaller intra-cluster distance than the left one. As discussed in the following Section \ref{sec:method}, intra-cluster distance directly reflects the possible loss of the FL. Hence, a much smaller intra-cluster distance indicates our proposed approach potentially reduces the loss of FL.

\subsection{Problem Formulation} \label{subsec:objective}

\paragraph{Solving a joint optimization on a distributed network.} The multi-center FL problem can be formulated as an optimisation framework as below.
\begin{align}\label{eq:mcfl} 
    \notag \min_{\{W_i\}, \{r_i^{(k)}\}, \{\tilde{W}^{(k)}\}}&\sum_{i=1}^{m} \alpha_i L_s(\mathcal{M}_i, \mathcal{D}_i, W_i)+\\
    &\dfrac{\lambda}{m} \sum_{k=1}^{K}  \sum_{i=1}^{m} r_i^{(k)} \distance(W_i, \tilde{W}^{(k)}),
\end{align}
where $\lambda$ controls the trade-off between supervised loss and distance.
We solve it by applying an alternative optimization between server and user: (1) on each node-$i$, we optimize the above objective w.r.t. $W_i$ while fixing all the other variables; and (2) on the server, we optimize $\{r_i^{(k)}\}, \{\tilde{W}^{(k)}\}$ for $i\in[m]$ and $k\in[K]$ while fixing all local models $\{W_i\}$.

\paragraph{Multi-center assignment at the server end.}
The second term in Eq.~\eqref{eq:mcfl} aims to minimize the distance between each local model and its nearest global model. 
Under the non-IID assumption, the data located at different devices can be grouped into multiple clusters where the on-device data in the same cluster are likely to be generated from one distribution. As illustrated on the right of Fig.~\ref{fig:aggregation}, we optimizes the assignments and global models by minimizing the intra-cluster distance, i.e,

\begin{align}\label{eq:new-objetive} 
    \min_{\{r_i^{(k)}\}, \{\tilde{W}^{(k)}\}}\dfrac{1}{m} \sum_{k=1}^{K}  \sum_{i=1}^{m} r_i^{(k)} \distance(W_i, \tilde{W}^{(k)}),
\end{align}
where cluster assignment $r_{i}^{(k)}$, as defined in Eq. \eqref{eq:cluster-assign}, indicates whether device-$i$ belongs to cluster-$k$, and $\tilde{W}^{(k)}$ is the parameters of the aggregated model for cluster-$k$.

\paragraph{Distance-constrained loss for local model optimization.} Because the distance between the local model and the global model are essential to our new loss, we don't expect the local model will be changed too much during the local updating stage. The new loss consists of a supervised learning loss and a regularization term to constrain the local model to ensure it is not too far from the global model. This kind of regularization term is also known as the proximal term in \cite{li2018federated} that can effectively limit the impact of the variable local updates in FL. 
We minimize the loss below for each local model $W_i$ as follows:

\begin{equation} \label{eq:local-model}
    \min_{W_i} \alpha_i L_s(\mathcal{M}_i, \mathcal{D}_i, W_i) + \dfrac{\lambda}{m} \sum_{k=1}^{K} r_i^{(k)}\distance(W_i, \tilde{W}^{(k)})
\end{equation}

\subsection{Optimization Algorithm}\label{subsec:algorithm}
In general, Expectation-Maximization (EM) \cite{bishop2006pattern} is an optimisation framework to solve alternative updating of multiple parameters, for example, K-Means clustering with updates on cluster centers and cluster assignments. However, in contrast to the general objective of clustering, our proposed objective, as described in Eq. \ref{eq:new-objetive}, has a dynamically changing $W_i$ during optimization.
Therefore, we adapt the Stochastic Expectation Maximization (SEM) \cite{cappe2009line} optimization framework by adding one step, i.e., updating $W_i$. In the modified SEM optimization framework, named federated SEM (FeSEM), we will iterative conduct a sequence of actions as below.

Firstly, for the \textbf{E-Step}, we update cluster assignment $r_i^{(k)}$ with fixed $W_i$. We calculate the distance between the cluster center and nodes -- each node is the model's parameters $W_i$, then update the cluster assignment $r_i^{(k)}$ by
\begin{equation} \label{eq:cluster-assign}
    r_i^{(k)} = 
    \left\{
        \begin{array}{ll}
        1, & if \; k = \argmin_j \distance(W_i, \tilde{W}^{(j)})  \\
        0, & otherwise.\\
        \end{array}
    \right.
\end{equation}

Secondly, for the \textbf{M-Step}, we update the cluster center $\tilde{W}^{(k)}$ according to the $W_i$ and $r_i^{(k)}$, i.e., 
\begin{equation} \label{eq:update-cluster}
    \tilde{W}^{(k)} = \frac{1}{\sum_{i=1}^m r_i^{(k)}} \sum_{i=1}^m r_i^{(k)} {W}_i.
\end{equation}

Thirdly, to \textbf{update the local models}, the global model's parameters $\tilde{W}^{(k)}$ are sent to each device in cluster $k$ to update its local model, and then we can fine-tune the local model's parameters ${W}_{i}$ using a supervised learning algorithm on its own private training data while considering the new loss as described in Eq. \ref{eq:local-model}. The local training procedure is a supervised learning task by adding a distance-based regularization term. The local model is initialized by the global model $\tilde{W}^{(k)}$ which belong to the cluster associated with the node.

Lastly, we repeat the three stochastic updating steps above until convergence. The sequential executions of the three updates comprise the iterations in FeSEM's optimization procedure. In particular, we sequentially update three variables $r_i^{(k)}$, $\tilde{W}^{(k)}$, and $W_i$ while fixing the other factors. These three variables are jointly used to calculate the objective of our proposed multi-center FL in Eq. \ref{eq:new-objetive}. 
\begin{algorithm}
\caption{FeSEM -- Federated Stochastic EM}
\label{alg:fesem}
\begin{algorithmic}[1]
    \State Initialize $K, \{W_i\}_{i=1}^m, \{\tilde{W}^{(k)}\}_{k=1}^K$
	\While{stop condition is not satisfied}
		\State \textbf{E-Step}:
		\State Calculate distance $d_{ik} \leftarrow \distance(W_i, \tilde{W}^{(k)})~~\forall i,k$ 
		\State Update cluster assignment $r_i^{(k)}$ using $d_{ik}$ (Eq. \ref{eq:cluster-assign})
		\State \textbf{M-Step}:
		\State Update $\tilde{W}^{(k)}$ using $r_i^{(k)}$ and $W_i$ (Eq. \ref{eq:update-cluster})
		\For{each cluster $k = 1, \dots K$ }
			\For{ $i \in  C_k$ }
				\State Send $\tilde{W}^{(k)}$ to device $i$
				\State $W_i \leftarrow \textbf{Local\_update}(i,\tilde{W}^{(k)})$ 
			\EndFor
		\EndFor
	\EndWhile
\end{algorithmic}
\end{algorithm}

\begin{algorithm}[t]
\caption{Local\_update}
\label{alg:local}
\begin{algorithmic}
	\State $i$ -- device index
	\State $\tilde{W}^{(k)}$ -- the model parameters from server
	\State $W_i$ -- updated local model
	\State Initialization: $W_i \leftarrow \tilde{W}^{(k)}$
	\For {$N$ local training steps}
		\State Update $W_i$ with training data $\mathcal{D}_i$ (Eq. \ref{eq:local-model})
	\EndFor
	\State Return $W_i$ to server\\
	
\end{algorithmic}
\end{algorithm}

We implement FeSEM in Algorithm \ref{alg:fesem} which is an iterative procedure. As elaborated in Section \ref{subsec:objective}, each iteration comprises of three steps to update three sets of variables that are cluster assignment $r$ for each client, the representation of cluster centre $\tilde{W}$, and the local models' parameters $W$, respectively. In particular, the update of the local model is executed on the client by fine-tuning the model as in Algorithm \ref{alg:local}.

\section{Some Possible Extensions}
To further handle heterogeneous data in FL scenario, our multi-center FL approach can be easily extended with other packages. We discuss two beneficial techniques here.
\subsection{Model Aggregation with Neuron Matching} \label{sec:neuron-matching}

The vanilla FL algorithm, FedAvg \cite{mcmahan2017communication}, uses model aggregation with index-based neuron matching which may cause the incorrect alignment. Neurons with similar functions usually take different indexes in two models. Recently, a function-based neuron matching \cite{Wang2020Federated} in FL is proposed to align two models by matching the neurons with similar functions.  In general, the index-based neuron matching can gradually align the neuron's functionality across nodes by repeatedly forcing each local model to be initialized using the same global model. However, the function-based neuron matching can speed up the convergence of neuron matching and preserve the unique functional neuron of the minority groups.

In this work, we integrate layer-wise matching and then averaging(MA) \cite{Wang2020Federated} into ours to increase the capacity to handle heterogeneous challenges. In the model aggregation step at the FL server, we need to align the layer-wise parameters among models from different clients. For example, given a layer with three neurons \{A1, A2, A3\} in model A, and a layer with neurons \{B1, B2, B3\} in model B, the neuron matching is to find an alignment between two sets that could be seen as a bi-party alignment problem. If we use index-based neuron matching, e.g. FedAvg \cite{mcmahan2017communication}, we just simply aggregate the parameters of A1 and B1 according to their index number in each model. If we choose function-based neuron matching, e.g. FedAvg \cite{Wang2020Federated}, we probably will aggregate the neuron A1 to B3 that they maybe share similar functions with different indexes in each model. The distance between the local model and the global model is the neuron matching score that is calculated by estimating the maximal posterior probability of the $j$-th client neuron $l$ generated from a Gaussian with mean $W_i$, and $\epsilon$ and $f(\cdot)$ are guided by the Indian Buffet Process prior \cite{yurochkin2019bayesian}.

\subsection{Selection of K}
The selection of $K$, the number of centers, is essential for a multi-center FL. In general, the $K$ is defined based on the prior experience or knowledge of data. If there is no prior knowledge, the most straightforward solution is to run the algorithm using different $K$ and then select the $K$ with the best performance in terms of accuracy or intra-cluster distance. Selecting the best $K$ in a large-scale FL system is time consuming, hence we simplify the process by running the algorithm on a small number of sampled nodes with several communication rounds. For example, we can randomly select 100 nodes and test $K$ in FL with three communication rounds only, and then apply the $K$ to the large-scale FL.

\begin{table}[t]\small
\centering
\begin{tabular}{lllll}
\hline
\textbf{DATASET}    & FEMNIST & FedCelebA \\ \hline
\# of data points   & 805,263 & 200,288   \\ 
\# of device        & 3,550   & 9,343     \\ 
\# of Classes       & 62      & 2         \\ 
Model architecture  & CNN     & CNN       \\ \hline
\end{tabular}
\caption{Statistics of datasets.
}
\label{table:dataset}
\end{table}

\section{Experiments}\label{sec:experiments}
As a proof-of-concept scenario to demonstrate the effectiveness of the proposed method, we experimentally evaluate and analyze FeSEM on two datasets.
\subsection{Training Setups}
\paragraph{Datasets.}
We employed two publicly-available  federated benchmarks datasets introduced in LEAF~\cite{caldas2018leaf}. LEAF is a benchmarking framework for learning in federated settings. 
The datasets used are Federeated Extended MNIST (FEMNIST)\footnote{\url{http://www.nist.gov/itl/products-and-services/emnist-dataset}}~\cite{cohen2017emnist} and Federated CelebA (FedCelebA)\footnote{\url{http://mmlab.ie.cuhk.edu.hk/projects/CelebA.html}}~\cite{liu2015deep}. 
We follow the setting of the benchmark data in LEAF. 
In FEMNIST, images is split according to the writers. 
For FedCelebA, images are extracted for each person and developed an on-device classifier to recognize whether the person smiles or not. 
A statistical description of the datasets is described in Table~\ref{table:dataset}.

\paragraph{Local model} 
We use a CNN with the same architecture from~\cite{liu2015deep}. Two data partition strategies are used: (a) an ideal IID data distribution using randomly shuffled data, (b) a non-IID partition by use a $ \mathbf{p}_{k} \sim Dir_{J}(0.5)$. Part of the code is adopted from \cite{Wang2020Federated}.
For FEMINST data, the local learning rate is 0.003 and epoch is 5. and for FedCelebA, 0.03 and 10 respectively.

\paragraph{Baselines.} In the scenario of solving statistical heterogeneity, we choose FL methods as follows:

\begin{enumerate}
    \setlength\itemsep{0em}
    \item \textbf{NonFed}: We will conduct the supervised learning task at each device without the FL framework.
    \item \textbf{FedSGD}: uses SGD to optimize the global model.
    \item \textbf{FedAvg}: is an SGD-based FL with weighted averaging. \cite{mcmahan2017communication} .
    \item \textbf{FedCluster}: is to enclose FedAvg into a hierarchical clustering framework \cite{sattler2019clustered}.
    \item \textbf{HypoCluster(K)}: is a hypothesis-based clustered-FL algorithm with different $K$ \cite{mansour2020three}.
    \item \textbf{Robust} our implementations based on the proposed method in \cite{ghosh2019robust}, see this baseline settings in Appendix.
    \item \textbf{FedDANE}: this is an FL framework with a Newton-type optimization method. \cite{li2019feddane}.
    \item \textbf{FedProx}: this is our our own implementations following \cite{li2018federated}. We set scaler of proximal term to 0.1. 
    \item \textbf{FedDist}: we adapt a distance based-objective function in Reptile meta-learning \cite{nichol2018reptile} to a federated setting.
    \item \textbf{FedDWS}: a variation of FedDist by changing the aggregation to weighted averaging where the weight depends on the data size of each device.
    \item \textbf{FeSEM($K$)}: our multi-center FL implemented on federated SEM with $K$ clusters. 
\end{enumerate}

\paragraph{Training settings.} We used 80\% of each device's data for training and 20\% for testing. 
For the initialization of the cluster centers in FeSEM, we conducted pure clustering 20 times with randomized initialization, and then the ``best'' initialization, which has the minimal intra-cluster distance, was selected as the initial centers for FeSEM. For the local update procedure of FeSEM, we set $N$ to 1, meaning we only updated $W_i$ once in each local update.

\paragraph{Evaluation metrics.} Given numerous devices, we evaluated the overall performance of the FL methods. We used classification accuracy and F1 score as the metrics for the two benchmarks. In addition, due to the multiple devices involved, we explored two ways to calculate the metrics, i.e., micro and macro. The only difference is that when computing an overall metric, ``micro'' calculates a weighted average of the metrics from devices where the weight is proportional to the data amount, while ``macro'' directly calculates an average over the metrics from devices.

\begin{table*}[htbp!]\small
    \begin{center}
      \begin{tabular}{|| l || c | c | c | c ||}
      \hline
        Dataset & \multicolumn{4}{|c|}{\textbf{FEMNIST}} \\\hline
        Metrics(\%) & Micro-Acc  & Micro-F1 & Macro-Acc & Macro-F1\\\hline \hline
        NoFed &  79.0$\pm 2.0$ &  67.6$\pm 0.6$ & 81.3$\pm 1.9$  &  51.0$\pm 1.2$     \\
        FedSGD & 70.1$\pm 2.2$  & 61.2$\pm 3.4$  & 71.5$\pm 1.8$  & 46.7$\pm 1.2$  \\
        FedAvg \cite{mcmahan2017communication} & 84.9$\pm 2.0$  & 67.9$\pm 0.4$  & 84.9$\pm 1.6$  &  45.4$\pm 1.9$       \\
        FedDist \cite{nichol2018reptile} &  79.3$\pm 0.8$ & 67.5$\pm 0.5$  &  79.8$\pm 1.1$ & 50.5$\pm 0.5$     \\
        FedDist+WS & 80.4$\pm 0.8$ &  67.2$\pm 1.6$ & 80.6$\pm 1.2$  & 51.7$\pm 1.1$    \\
        Robust(TKM) \cite{ghosh2019robust} & 78.4$\pm 1.0$  & 53.1$\pm 0.5$  & 77.6$\pm 0.7$ & 53.6$\pm 0.7$     \\
        FedCluster \cite{sattler2019clustered} &  84.1$\pm 1.1$&  64.3$\pm 1.3$ & 84.2$\pm 1.0$ & \textbf{64.4}$\pm 1.6$   \\
        HypoCluster(3) \cite{mansour2020three} &  82.5$\pm 1.7$ &  61.3$\pm 0.6$ & 82.2$\pm 1.3$ & 61.6$\pm 0.9$  \\
        FedDane \cite{li2019feddane} & 40.0$\pm 2.9$& 31.8$\pm 3.1$  & 41.7$\pm 2.4$ & 31.7$\pm 1.6$     \\
        FedProx \cite{li2018federated} & 72.6$\pm 1.8$ &  62.8$\pm 1.6$ & 74.3$\pm 2.1$  & 50.6$\pm 1.2$    \\ \hline
        FeSEM(2) &   84.8$\pm 1.1$& 65.5$\pm 0.4$  & 84.8$\pm 1.6$  & 52.0$\pm 0.5$   \\
        FeSEM(3) &  87.0$\pm 1.2$& 68.5$\pm 2.0$  & 86.9$\pm 1.2$  & 41.7$\pm 1.5$   \\
        FeSEM(4) &  \textbf{90.3}$\pm 1.5$ & 70.6$\pm 0.9$  & \textbf{91.0}$\pm 1.8$  & 53.4$\pm 0.6$   \\
        FeSEM-MA(3)      & \textbf{90.4}$\pm 1.5$ & \textbf{71.4}$\pm 0.5$ & 87.0$\pm 2.0$ & 64.3$\pm 0.5$   \\
        \hline
      \end{tabular}
    \end{center}
    \caption{Comparison of our proposed FeSEM($K$) algorithm with the baselines on FEMNIST. Note the number in parenthesis following ``FeSEM'' denotes the number of clusters, $K$. }
    \label{table:comparison-1}
\end{table*}

\begin{table*}[htbp!]\small
    \begin{center}
      \begin{tabular}{|| l || c | c | c | c ||}
      \hline
        Dataset & \multicolumn{4}{|c||}{\textbf{FedCelebA}} \\\hline
        Metrics(\%) & Micro-Acc  & Micro-F1 & Macro-Acc & Macro-F1\\\hline \hline
        NoFed &  83.8$\pm 1.4$ &  66.0$\pm 0.4$ & 83.9$\pm 1.6$ & 67.2$\pm 0.6$     \\
        FedSGD  &  75.7$\pm 2.3$ & 60.7$\pm 2.4$  &  75.6$\pm 2.0$ & 55.6$\pm 2.6$     \\
        FedAvg \cite{mcmahan2017communication}  &  86.9$\pm 0.5$ &  \textbf{78.0}$\pm 1.0$ & 86.1$\pm 0.4$  & 54.2$\pm 0.6$      \\
        FedDist \cite{nichol2018reptile} &  71.8$\pm 0.9$& 61.0$\pm 0.8$  & 71.6$\pm 1.0$  &  61.1$\pm 0.7$    \\
        FedDist+WS & 73.4$\pm 1.7$ & 59.3$\pm 0.9$  & 73.4$\pm 1.9$ &  50.3$\pm 0.5$    \\
        Robust(TKM) \cite{ghosh2019robust} &  90.1$\pm 1.3$&  68.0$\pm 0.7$ & 90.1$\pm 1.3$ &  68.3$\pm 1.1$    \\
        FedCluster \cite{sattler2019clustered} &  86.7$\pm 0.7$ & 67.8$\pm 0.9$  & 87.0$\pm 0.9$ &  67.8$\pm 1.3$   \\
        HypoCluster(3) \cite{mansour2020three} &  76.1$\pm 1.5$ & 53.5$\pm 1.0$  & 72.7$\pm 1.8$ &  53.8$\pm 1.9$   \\
        FedDane \cite{li2019feddane} & 76.6$\pm 1.1$ &  61.8$\pm 2.0$ &  75.9$\pm 1.0$ & 62.1$\pm 2.2$     \\
        FedProx \cite{li2018federated}  &  83.8$\pm 2.0$& 60.9$\pm 1.2$  &  84.9$\pm 1.8$&  65.7$\pm 1.2$    \\ \hline
        FeSEM(2) & 89.1$\pm 1.3$  & 64.6$\pm 1.0$  & 89.0 $\pm 1.3$  &  56.0$\pm 1.3$ \\
        FeSEM(3) & 88.1$\pm 1.9$  & 64.3$\pm 0.8$  & 87.5$\pm 2.0$  &  55.9$\pm 0.8$  \\
        FeSEM(4) & \textbf{93.6}$\pm 2.7$  & \textbf{74.8}$\pm 1.5$  & \textbf{94.1}$\pm 2.2$  &  \textbf{69.5}$\pm 1.1$  \\
        FeSEM-MA(3) & 84.5$\pm 0.8$ & 64.1$\pm 0.7$ & 85.1$\pm 1.0$  & 63.0$\pm 1.3$ \\
        \hline
      \end{tabular}
    \end{center}
    \caption{Comparison of our proposed FeSEM($K$) algorithm with the baselines on FedCelebA. Note the number in parenthesis following ``FeSEM'' denotes the number of clusters, $K$. }
    \label{table:comparison-2}
\end{table*}

\subsection{Experimental Study}

\paragraph{Comparison study}
As shown in Table~\ref{table:comparison-1} and \ref{table:comparison-2}, we compared our proposed FeSEM with the baselines and found that FeSEM achieves the best performance in most cases. But, it is observed that the proposed model achieves an inferior performance for Micro F1 score on the FedCelebA dataset. A possible reason for this is that our objective function defined in Eq.~\ref{eq:new-objetive} does not take into account the number of samples per device as important weights. Hence, our model is able to deliver a significant improvement in terms of ``macro'' metrics. 
Furthermore, as shown in the last four rows in Table~\ref{table:comparison-1} and \ref{table:comparison-2}, we found that FeSEM with a larger number of clusters empirically achieves a better performance, which verifies the correctness of the non-IID assumption of the data distribution.
It is worth noting that the FeSEM-MA is comparable or worsen than FeSEM in our experiment. Because the Matched Averaging method usually works well that each client is randomly initialised while the FedAvg and our FeSEM initialized all clients using the global model. Different initialization strategies will cause different scales of diversity across local models, thus contributing to the performance in different ways.

We also notice that FedAvg and NoFed perform better than some clustered FL baseline. The reason is that our experiment using the original federated learning setting that only split the dataset according to users or by the random split. Therefore, the non-IID and clustering assumption is not obvious in this situation. Therefore, the FedAvg and noFed perform well. In recent, some advanced data split methods have been used to further disturb the non-IID distribution across clients. For example, exaggerate the label distribution across devices by aggregating sample imbalance, or randomly assign different label set to each device. In these new settings, the FedAvg will perform worse, and the clustered FL methods will perform relatively better. Due to the limited space, we will use the original FL setting, and leave the new setting for future work.

\begin{figure}[t] %
	\centering
	\includegraphics[width=0.75\textwidth]{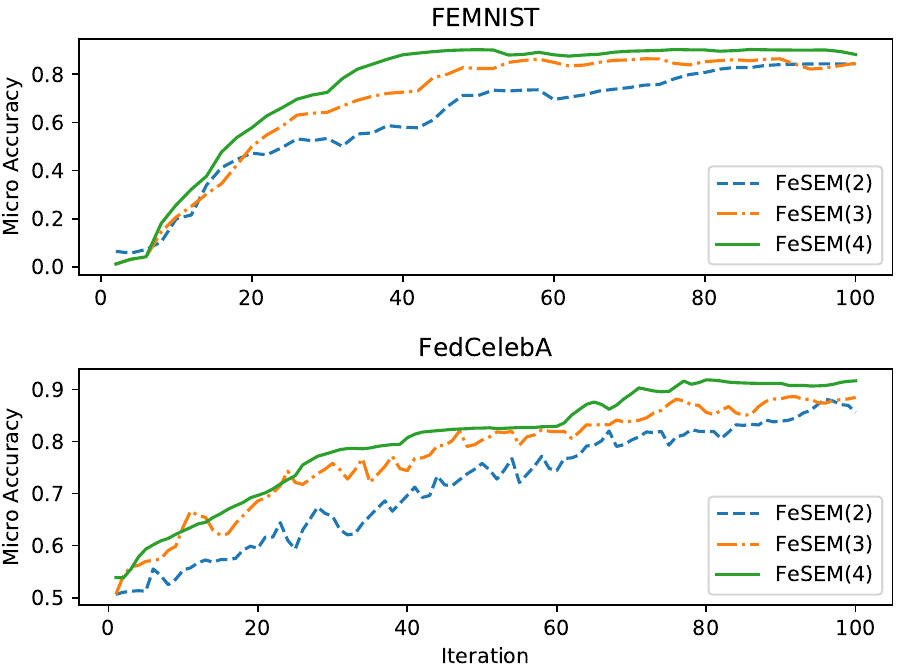}
	\caption{\small Convergence analysis for the proposed FeSEM with different cluster number (in parenthesis) in terms of micro-accuracy.}
	\label{fig:convergence} 
	\centering
\end{figure}

\paragraph{Convergence}
To verify the convergence of the proposed approach, we conducted a convergence analysis by running FeSEM with different cluster numbers $K$ (from 2 to 4) in 100 iterations. As shown in Fig.~\ref{fig:convergence}, FeSEM can efficiently converge on both datasets and it can achieve the best performance with the cluster number $K=4$.

\begin{figure}[t] 
    \centering
    \includegraphics[width=0.98\textwidth]{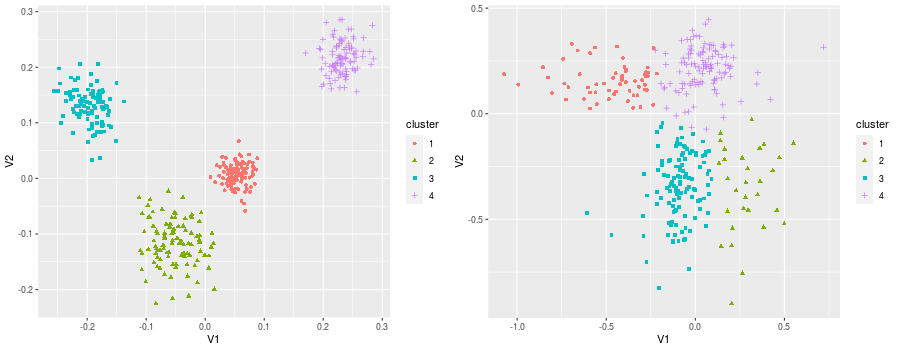}
    \caption{\small Clustering analysis for different local models (using PCA) derived from FeSEM(4) using FEMNIST and Celeba data.}
    \label{fig:model-clustering}
\end{figure}

\paragraph{Clustering analysis}
To check the effectiveness of our proposed optimization method and whether the devices grouped into one cluster have similar model, we conducted a clustering analysis via an illustration. 
We used two-dimensional figures to display the clustering results of the local models derived from FeSEM(4) on the FEMNIST dataset. In particular, we randomly chose 400 devices from the dataset and plotted each device's local model as one point in the 2D space after PCA dimension reduction. As shown in Fig.~\ref{fig:model-clustering}, the dataset suitable for four clusters that are distinguishable to each other. 
It is worth noting that the clustering algorithm will converge very fast. In general, it takes no more than 10 iterations to converge. One can find detailed theoretical and empirical analyses on the convergence for general clustered FL in a recent work~\cite{ma2022convergence}.

\paragraph{Case study on clustering} 
To intuitively judge whether nodes grouped into the same cluster have a similar data distribution, we conducted case studies on a case of two clusters that are extracted from a trained FeSEM(2) model. 
For FMNIST, as shown on the top of Fig.~\ref{fig:case_study}, cluster on the right consists writers who are likely to recognize hand-writings with a smaller font, and on the left consists writers who are likely to recognize hand-writing with a bolder and darker font. For FedCelebA, see full face images in Appendix section 2, the face recognition task in cluster1 is likely to handle the smiling faces with a relatively simple background, also exhibits to be young people. While cluster on the right is likely to handle the faces with more diverse background and also seems to be more older people.

\section{Case study}
\begin{figure}[thbp!] %
    \setlength{\belowcaptionskip}{-10pt}
    \centering
    \includegraphics[width=0.49\textwidth]{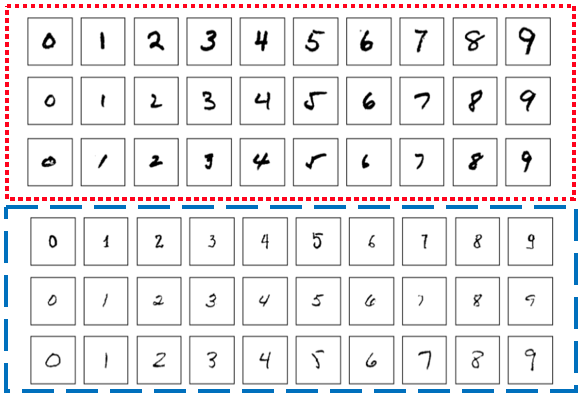}
    \includegraphics[width=0.49\textwidth]{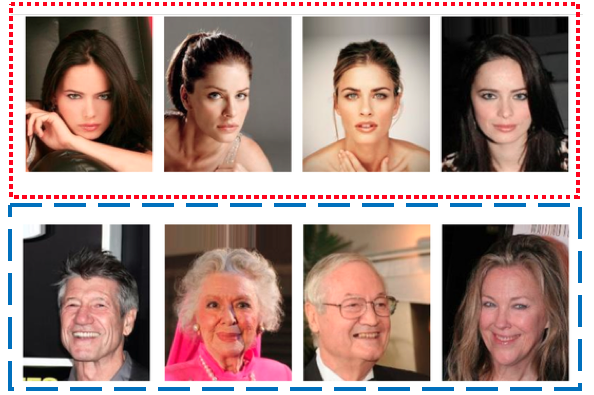}
    \caption{Cluster demonstrations on FEMINIST (left) and CIFAR-10 (right). Two rectangles represent different cluster discovered by the proposed method.}
    \label{fig:case_study}
\end{figure}

The cluster results are based on the distance metric for client-wise model parameters. The model could be viewed as a map function $f: x \rightarrow y$ that is to map the input $x$ into an output $y$. The function is parameterised by model parameters $w$, thus a similar $w$ pair indicate that the two clients are likely to have similar preferences in decision-making. 
As shown in Figure \ref{fig:case_study}, there are two clusters for FEMNIST and CIFAR-10 datasets respectively. The left part shows the clustering effect of FeSEM on dataset MINIST by writers, on the red rectangle (upper) are three writers handwritten digits which are smaller and lighter than those in the blue rectangle (bottom). The right figure shows the clustering of CIFAR-10 data with 10 classes in which one class is about people. In the class of people, our algorithm finds one cluster to represent young beauties as shown in the red rectangle (upper) and another cluster to represent the aging people in the blue rectangle (bottom).

\section{Conclusion}\label{Conc}
This work proposes a novel FL algorithm to tackle the non-IID challenge, which is towards better personalization of  decision-making in non-IID FL. Since the proposed method uses Kmeans as the clustering algorithm, it also suffers the drawbacks of Kmeans, e.g., computational efficiency for high dimension data and robustness to outliers. Moreover, the clustering algorithm usually requires the full participation of all clients in the training process. A more practical clustered federated learning could be developed to address the large-scale federated learning system, and also to explore how to adapt to new clients out of the training set.

\section{Conflict of interest}
All authors declare that they have no conflicts of interest.

\backmatter

\bibliography{references}


\end{document}